\documentclass[11pt,a4paper]{article}
\usepackage[hyperref]{eacl2021}
\usepackage{times}
\usepackage{latexsym}

\usepackage{microtype}
\usepackage{bm}
\usepackage{multicol}
\usepackage{multirow}
\usepackage{booktabs}
\usepackage{natbib}
\usepackage{amsthm}
\usepackage{amssymb}
\usepackage{amsmath}
\usepackage{amsfonts}
\usepackage{xcolor}
\usepackage{microtype}
\usepackage{comment}
\usepackage{caption,subcaption}
\usepackage{tabularx}
\usepackage{colortbl}
\usepackage{url}
\usepackage{arydshln}
\usepackage{graphicx}

\aclfinalcopy 

\title{\textsc{DebIE}: A Platform for Implicit and Explicit Debiasing \\ of Word Embedding Spaces}

\author{Niklas Friedrich, Anne Lauscher, Simone Paolo Ponzetto and Goran Glava\v{s} \\
  Data and Web Science Group \\
  School of Business Informatics and Mathematics \\
  University of Mannheim \\ 
  \texttt{nfriedri@mail.uni-mannheim.de} \\
  \texttt{\{anne,simone,goran\}@informatik.uni-mannheim.de}}

\date{}

\begin{document}
\maketitle

\begin{abstract}
Recent research efforts in NLP have demonstrated that distributional word vector spaces often encode stereotypical human biases, such as racism and sexism. With word representations ubiquitously used in NLP models and pipelines, this raises ethical issues and jeopardizes the fairness of language technologies. While there exists a large body of work on bias measures and debiasing methods, to date, there is no platform that would unify these research efforts and make bias measuring and debiasing of representation spaces widely accessible. 
In this work, we present \textsc{DebIE}, the first integrated platform for (1) measuring and (2) mitigating bias in word embeddings. Given an (i) embedding space (users can choose between the predefined spaces or upload their own) and (ii) a bias specification (users can choose between existing bias specifications or create their own), \textsc{DebIE} can (1) compute several measures of implicit and explicit bias and modify the embedding space by executing two (mutually composable) debiasing models. \textsc{DebIE}'s functionality can be accessed through four different interfaces: (a) a web application, (b) a desktop application, (c) a REST-ful API, and (d) as a command-line application.\footnote{Videos demonstrating the usage of the \textsc{DebIE} application and command-line tool are available at \url{https://tinyurl.com/y2ymujus}} \textsc{DebIE} is available at: \url{debie.informatik.uni-mannheim.de}

\end{abstract}

\section{Introduction}
\label{s:intro}
 Ethical and fair natural language processing is an essential precondition for widespread societal adoption of language technologies. In recent years, however, distributional language representations built from large corpora have been shown to encode human-like biases, like racism and sexism \cite[\textit{inter alia}]{Bolukbasi2016,zhao2019gender,Lauscher2020,nadeem2020stereoset}. 
 At the word level, most embedding spaces, across a range of embedding models and languages \cite{Lauscher2019are}, encode human biases that can be exemplified in biased analogies, such as the famous example of sexism: $\overrightarrow{\text{man}} - \overrightarrow{\text{programmer}} \approx \overrightarrow{\text{woman}} - \overrightarrow{\text{homemaker}}$ \cite{Bolukbasi2016}. While this is not surprising, given the distributional nature of word representation models \cite{Harris:1954} 
 it is -- depending on the sociotechnical context -- an undesired artefact of distributional representation learning \cite{blodgett2020language} which can, in turn, lead to unfair decisions in downstream applications. A number of different measures for quantifying biases in representation spaces have been proposed in recent years \cite{Caliskan2017,Gonen2019,Dev2019,GargE3635,Lauscher2020} and even more models for removing or attenuating such biases have been developed \cite[\textit{inter alia}]{zhao2019gender,bordia2019identifying,dinan-etal-2020-queens,webster2020measuring,qian2019reducing}. What is still missing, however, is the ability to seamlessly apply different bias measures and debiasing models on arbitrary embedding spaces and for custom (i.e., user-specified) bias specifications.      

\vspace{-0.5em}
 
In this work, we address this gap by introducing \textsc{DebIE}, the first integrated platform offering bias measurement and mitigation for arbitrary static embedding spaces and bias specifications. The \textsc{DebIE} platform is grounded in the general framework for \emph{implicit} and \emph{explicit} debiasing of word embedding spaces \cite{Lauscher2020}. Within this framework, an implicit bias consists of measurable discrepancies between two target term sets, which can, for instance, describe a dominant and a minoritized social group~\citep{d2020data}. In contrast, an explicit bias is a bias between such target term sets towards certain attribute terms groups. Our platform allows for both implicit and explicit bias specifications, incorporating a range of different measures for quantifying embedding space bias \cite{Caliskan2017,Gonen2019,Dev2019} and a pair of mutually composable methods for bias mitigation. \textsc{DebIE}'s functionality for measuring and mitigating biases in distributional word vector spaces is accessible via four different interfaces: as a web application, desktop application, via a RESTful application programming interface (API), and as a command-line tool. 
We believe that \textit{DebIE} will, by offering to test arbitrary embedding spaces for custom user-defined biases, stimulate a wider exploration of the presence of a broader set of human biases in distributional representation spaces.



\section{Related Work}
First, we describe related research on bias evaluation and debiasing and then turn our attention to existing bias mitigation platforms.

\paragraph{Bias Measures and Mitigation Methods.} There is an extensive body of research on bias detection and bias mitigation in natural language processing. Due to space limitations, here we only provide a brief overview and refer the reader to a recent survey of the field for more information \cite{blodgett2020language}. \newcite{Bolukbasi2016} were the first to show stereotypical bias to exist in word embedding models and proposed \emph{hard debiasing}, the first word embedding bias mitigation algorithm. Subsequently, \citet{Caliskan2017} introduced the well-known Word Embedding Association Test (WEAT), inspired by the Implicit Association Test~\citep{Nosek02harvestingimplicit}, which measures biased associations in human subjects in terms of response times when exposed to sets of stimuli. WEAT, in turn, reflects the strength of associations in terms of semantic similarity between word vectors. \newcite{mccurdy2018} study gender bias with WEAT in three other languages (Dutch, German, and Spanish). Extending upon this, \newcite{Lauscher2019are} translated the WEAT tests to $6$ more languages (German, Spanish, Italian, Russian, Croatian, Turkish), allowing for multilingual and cross-lingual analysis of biases captured by the specifications of the original WEAT. They later extended the set of supported languages with Arabic \citep{lauscher-etal-2020-araweat}. 

\citet{Dev2019} proposed a linear projection model for debiasing along with two bias evaluation measures: the Embedding Coherence Test (ECT) and the Embedding Quality Test (EQT) and propose methods for removing the (explicit) bias based on computing the direction vector of the bias. While their method successfully removes the \emph{explicit} bias, i.e., bias between sets of \textit{target} terms (e.g., male terms like \textit{man}, \textit{father}, and \textit{boy} vs. female terms like \textit{woman}, \textit{mother}, and \textit{girl}) \emph{with respect to} sets of \textit{attribute} terms (e.g., profession terms, such as \textit{scientist} or \textit{artist}), \citet{Gonen2019} show that \emph{implicit} bias between the sets of target terms remains even after (explicit debiasing) and that the terms from one target set are still clearly discernible from the terms of the other set in the embedding space. Based on this finding, \citet{Lauscher2020} systematized the preceding work and proposed a general framework for bias measurement and debiasing, encompassing a range of existing and newly proposed measures and mitigation methods, which operate either on \emph{explicit} or \emph{implicit} bias specifications. Their framework arguably allows for a more holistic assessment of bias in word vector spaces and ensures interoperability between bias mitigation models and bias specifications. Our \textsc{DebIE} platform makes this holistic framework for measuring and mitigating biases widely accessible and applicable (1) for arbitrary user-defined bias specification to (2) arbitrary pretrained word embedding spaces.

\paragraph{Bias mitigation platforms.} The landscape of the \textit{off-the-shelf} solutions for measuring and mitigating bias for machine learning applications is extremely scarce. To the best of our knowledge, the only such tool is \emph{AI Fairness 360}~\cite{AIF360}, an extensible open-source toolkit which offers a set of algorithms for detecting and mitigating unwanted bias in datasets and machine learning models. 
It addresses bias by integrating fairness algorithms along the machine learning pipeline, i.e., fair pre-processing, fair in-processing, and fair postprocessing. In contrast, \textsc{DebIE} specifically targets biases in distributional word vector spaces (as an ubiquitous component of modern NLP pipelines) by integrating a series of word embedding bias tests and mitigation algorithms not covered by more general tools like AI Fairness 360. 

\section{\textsc{DebIE}: System Description}
We first explain the two types of bias specifications support by \textsc{DebIE} (\emph{implicit} and \emph{explicit}), then proceed to describe the concrete bias specifications and debiasing algorithms bundles in the system. Finally, we provide details of \textsc{DebIE}'s architecture and interfaces through which the bias measuring and mitigation functionality can be accessed. All code is publicly available on GitHub.\footnote{\url{https://github.com/umanlp/debie-frontend} \\ \url{https://github.com/umanlp/debie-backend}}

\subsection{Implicit and Explicit Bias Specifications}
\textsc{DebIE} supports measuring of \textit{implicit} or \textit{explicit} biases for a given word embedding space and, respectively, implicit or explicit \textit{debiasing} of the given space. Both implicit bias specifications $B_I$ and explicit bias specifications $B_E$ specify two sets of \textit{target} terms, $T_1$ and $T_2$ that capture the dimension of the bias. For example, if measuring a \textit{gender} bias, $T_1$ would contain male terms (e.g., \textit{man}, \textit{father}) and $T_2$ female terms (e.g., \textit{woman}, \textit{girl}, \textit{grandma}).\footnote{The bias measures implemented in \textsc{DebIE} do not require the terms between the target lists to be paired. Accordingly, the two lists also do not need to be of the same length.} While an implicit bias specification is fully specified with the two target lists, $B_I = (T_1, T_2)$, an explicit specification additionally requires two sets of attributes $A_1$ and $A_2$, $B_E = (T_1, T_2, A_1, A_2)$, capturing the groups of terms towards which the target groups are expected to exhibit significantly different level of association. For example, for a gender bias, one would expect male terms to be more strongly associated with career terms (e.g., $A_1$ could contain terms like \textit{programmer}), whereas female terms could be closer to family-related terms (e.g., $A_2$ could contain terms like \textit{homemaker}). The input for \textsc{DebIE} consists of an embedding space $\mathbf{X} \in \mathbb{R}^d$ and a bias specification, (implicit or explicit). Explicit debiasing methods (i.e., methods that operate on explicit bias specifications) cannot be executed when the provided bias specification is implicit ($B_I$).\footnote{Conversely, implicit debiasing methods, i.e., ones that require only $T_1$ and $T_2$, can be applied if an explicit specification is provided. In that case, we simply convert $B_E$ to $B_I$ by discarding the provided attribute sets.}       


\subsection{Bias Measures}
\label{sec:measures}

\textsc{DebIE} provides three measures that capture explicit bias (i.e., apply only if an explicit bias specification is provided), and two tests that measure implicit bias. Because debiasing methods (see \S\ref{sec:debiasing}) make perturbations to the embedding space, we additionally couple the bias tests with measures of semantic quality of the distributional space.

\vspace{0.5em}

\noindent\textbf{Word Embedding Association Test (WEAT).} Given an explicit bias test specification $B_E=(T_1,T_2,A_1,A_2)$, WEAT~\citep{Caliskan2017} computes the effect size quantifying the amount of bias as follows:


\small{
\begin{equation*}
    s(T_1, T_2, A_1, A_2) = \hspace{-0.5em}\sum_{t_1 \in T_1}{\hspace{-0.3em}s(t_1, A_1, A_2)} - \hspace{-0.5em}\sum_{t_2 \in T_2}{\hspace{-0.3em}s(t_2, A_1, A_2)}, 
\end{equation*}}

\vspace{-1em}

\normalsize
\noindent with associative difference of term $t$ given as: 


\small{
\begin{equation*}
    s(t, A_1, A_2) \hspace{-0.1em}=\hspace{-0.1em} \frac{1}{|A_1|}\sum_{a_1 \in A_1}{\hspace{-0.5em}\textnormal{cos}(\mathbf{t}, \mathbf{a_1})} - \frac{1}{|A_2|}\sum_{a_2 \in A_2}{\hspace{-0.5em}\textnormal{cos}(\mathbf{t}, \mathbf{a_2})}, 
\end{equation*}}
\normalsize
\noindent with $\mathbf{t}$ as the word embedding of the target term $t$ and $\mathit{cos}$ as the cosine of the angle between the two vectors.
To estimate the significance of the effect size, we follow \citet{Caliskan2017} and compute the non-parametric permutation test in which the $s(T_1, T_2, A_1, A_2)$ is compared to $s(X_1, X_2, A_1, A_2)$, where ($X_1$, $X_2$) denotes a random, equally-sized split of terms from $T_1 \cup T_2$. 

\vspace{0.5em}
\noindent\textbf{Embedding Coherence Test (ECT).} Given an explicit bias specification with a single attribute set $B_E = (T_1, T_2, A)$ with $A=A_1\cup A_2$, ECT~\citep{Dev2019} quantifies the presence of the bias as the (lack of) correlation of the distances of the mean vectors of the target term sets $T_1$ and $T2$ with the attribute terms in $A$. The lower the correlation, the higher the bias. To this end, we compute the mean vectors $\mathbf{t_1}$ and $\mathbf{t_2}$ as averages of the vector representations of the terms in $T_1$ and $T_2$. Next, we compute two vectors containing the cosine similarities of each of the terms in $A$ with $\mathbf{t_1}$, as well as with $\mathbf{t_2}$, respectively. The final score is Spearman's rank correlation coefficient of the obtained vectors of cosine similarity scores.  


\vspace{0.5em}
\noindent\textbf{Bias Analogy Test (BAT).} BAT~\citep{Lauscher2020} assesses the amount of biased analogies that can be retrieved from an embedding space based on the explicit bias specification $B_E = (T_1, T_2, A_1, A_2)$. We first create all possible biased analogies from $B_E$:  $\mathbf{t}_1 - \mathbf{t}_2 \approx \mathbf{a}_1 - \mathbf{a}_2$ for $(t_1, t_2, a_1, a_2) \in T_1 \times T_2 \times A_1 \times A_2$. Next, from each of these analogies, two query vectors are computed: $\mathbf{q}_1 = \mathbf{t}_1 - \mathbf{t}_2 + \mathbf{a}_2$ and $\mathbf{q}_2 = \mathbf{a}_1 - \mathbf{t}_1 + \mathbf{t}_2$ for each 4-tuple $(t_1, t_2, a_1, a_2)$. We then rank all attribute vectors in $\mathbf{X}$ according to the Euclidean distance to the query vector.  We report the percentage of cases in which: (1) $a_1$ is ranked higher than a term $a'_2 \in A_2\setminus\{a_2\}$ for $\mathbf{q}_1$ and (2) $a_2$ is ranked higher than a term $a'_1 \in A_1\setminus\{a_1\}$ for $\mathbf{q}_2$. 


\vspace{0.5em}
\noindent\textbf{Implicit Bias Tests (IBT).} As proposed by \citet{Gonen2019}, the amount of implicit bias corresponds to the accuracy with which two target term sets can be separated. We report the score of two methods: (1) clustering accuracy with K-Means++~\cite{Arthur2007kmeans}, and (2) classification accuracy based on Support Vector Machines with Gaussian kernel. We carry out the latter via leave-one-out cross-validation (i.e., we train on all words from both target lists, leaving one term for prediction). 

\vspace{0.5em}
\noindent\textbf{Semantic Quality Tests (SQ). } The debiasing models (\ref{sec:debiasing}) modify the embedding space. While they reduce the bias, they may reduce the general semantic quality of the embedding space, which could be detrimental for model performance in downstream applications. This is why we couple the bias tests with measures of semantic word similarity on two established word-similarity datasets: SimLex-999~\citep{hilldoi:10.1162} or WordSim-353 \cite{finkelstein2001placing}. We compute the Spearman correlation between the human similarity scores assigned to word pairs and corresponding cosines computed from the embedding space.

\subsection{Debiasing Methods}
\label{sec:debiasing}

\textsc{DebIE} encompasses implementations of two debiasing models from \cite{Lauscher2020}, for which an implicit bias specification suffices:\footnote{Note that any explicit bias specification is trivially reduced to an implicit one by discarding the attribute term sets.}
%
%

\vspace{0.5em}
\noindent\textbf{General Bias Direction Debiasing (GBDD).}
As an extension of the linear projection model of \newcite{Dev2019}, GBDD relies on identifying the bias direction in the distributional space.
Let $(t^i_1, t^j_2)$ be word pairs with $t^i_1 \in T_1$, $t^j_2 \in T_2$, respectively. First, we obtain partial bias direction vectors $\mathbf{b_{ij}}$
by computing the difference between the respective vectors for each pair $\bm{b_{ij}} = \bm{t^i_1}-\bm{t^j_2}$. We then stack all partial direction vector, obtaining the bias matrix $\bm{B}$. The global bias direction vector $\bm{b}$ then corresponds to the top singular value of $\bm{B}$, i.e., the first row of matrix $\bm{V}$, with $\bm{U\Sigma V^\top}$ as the singular value decomposition of $\bm{B}$. 
We then obtain the debiased version of the space $\bm{X}$ as:

\vspace{-1em}

{
\begin{equation*}
    \text{GBDD}(\mathbf{X}) = \mathbf{X} - \langle \mathbf{X}, \mathbf{b} \rangle \mathbf{b},
\end{equation*}}

\vspace{-1.5em}

\noindent with $\langle \bm{X}, \bm{b} \rangle$ denoting dot products between rows of $\bm{X}$ and $\bm{b}$. As such, the closer the word embedding is to the bias direction, the more it gets corrected. 

\vspace{0.5em}
\noindent\textbf{Bias Alignment Model (BAM).} Inspired by previous work on projection-based cross-lingual word embedding spaces~\cite{smith2017offline, glavas2019properly}, BAM focuses on \emph{implicit} debiasing by treating the target term sets $T_1$, and $T_2$ of an implicit bias specification $B_I$ as ``translations'' of each other and learning the linear projection of the embedding spaces w.r.t. itself \cite{Lauscher2020}. 
First, we build all possible word pairs $(t^i_1, t^j_2)$, $t^i_1 \in T_1$, $t^j_2 \in T_2$ and stack the respective word vectors of the left and right pairs to obtain matrices  $\bm{X}_{T_1}$ and $\bm{X}_{T_2}$. We then learn the orthogonal mapping matrix $\bm{W_X} = \bm{U}\bm{V}^\top$, with $\bm{U}\bm{\Sigma}\bm{V}^\top$ as the singular value decomposition of $\bm{X}_{T_2}\bm{X}_{T_1}^\top$. 
In the last step, the original space and its ``translation'' $\bm{X}= \bm{X}\bm{W_X}$ (which is equally biased), are averaged to obtain the debiased embedding space:

\vspace{-0.5em}

\begin{equation*}
    \text{BAM}(\bm{X}) = \frac{1}{2} (\bm{X} + \bm{X}\bm{W_X})\,.
\end{equation*}


\noindent Note that \textsc{DebIE} can trivially compose the two debiasing models -- the resulting space after applying GBDD (BAM) can be the input for BAM (GBDD). 


\subsection{Integrated Data} 
\label{sec:data}

%
%
\setlength{\tabcolsep}{3pt}
\begin{table*}[t]
\centering
{\scriptsize
\begin{tabular}{c | lllll}
\toprule
\textbf{Test} & \textbf{Type} & \textbf{Target Set \#1} & \textbf{Target Set \#2} & \textbf{Attribute Set \#1} & \textbf{Attribute Set \#2} \\ \midrule
1 & Universal & Flowers (e.g.,~\emph{aster}, \emph{tulip})
& Insects (e.g.,~\emph{ant}, \emph{flea}) & Pleasant (e.g.,~\emph{health}, \emph{love}) & Unpleasant~(e.g., \emph{abuse})
\\
2 & Militant &  Instruments (e.g.,~\emph{cello}, \emph{guitar}) 
& Weapons (e.g.,~\emph{gun}, \emph{sword}) &
Pleasant & Unpleasant\\
3 & Racist & Euro-American names (e.g.,~\emph{Adam})
& Afro-American names (e.g.,~\emph{Jamel})
& Pleasant (e.g.,~\emph{caress}) & Unpleasant (e.g.,~\emph{abuse})\\
4 & Racist & Euro-American names (e.g.,~\emph{Brad})
& Afro-American names (e.g.,~\emph{Hakim})
& Pleasant 
& Unpleasant 
\\
5 & Racist & Euro-American names
& Afro-American names
& Pleasant  (e.g.,~\emph{joy}) & Unpleasant  (e.g.,~\emph{agony})\\
6 & Gender & Male names (e.g.,~\emph{John}) & Female names (e.g.,~\emph{Lisa}) & Career (e.g. \emph{management})
& Family (e.g.,~\emph{children}) \\
7 & Gender & Math (e.g.,~\emph{algebra}, \emph{geometry}) & Arts (e.g.,~\emph{poetry}, \emph{dance}) & Male (e.g.,~\emph{brother}, \emph{son}) & Female (e.g.,~\emph{woman}, \emph{sister}) \\
8 & Gender & Science (e.g.,~\emph{experiment}) & Arts & Male & Female\\
9 & Disease & Physical condition (e.g.,~\emph{virus}) & Mental condition (e.g.,~\emph{sad}) & Long-term (e.g.,~\emph{always}) & Short-term (e.g.,~\emph{occasional})\\
10 & Age & Older names (e.g.,~\emph{Gertrude}) & Younger names (e.g.,~\emph{Michelle}) & Pleasant & Unpleasant \\ 
\bottomrule
\end{tabular}
}
\caption{WEAT bias test specifications provided by \textsc{DebIE}.}
\label{tbl:weat}
\end{table*}
\textsc{DebIE} is designed as a general tool, which allows user to upload their own embedding spaces and define their own bias specifications for testing and/or debiasing. Nonetheless, we include into the platform a set of commonly used bias specifications and word embedding spaces.
Concretely, \textsc{DebIE} includes the whole WEAT test collection~\citep{Caliskan2017}, containing the explicit bias specifications summarized in Table~\ref{tbl:weat}. \textsc{DebIE} also comes with three word embedding spaces, pretrained with different models: 
(1) fastText~\cite{Bojanowski:2017tacl},\footnote{\url{https://dl.fbaipublicfiles.com/fasttext/vectors-wiki/wiki.en.vec}} (2) GloVe~\cite{pennington2014glove},\footnote{\url{http://nlp.stanford.edu/data/glove.6B.zip}} and (3) CBOW~\cite{mikolov2013distributed}.\footnote{\url{https://drive.google.com/file/d/0B7XkCwpI5KDYNlNUTTlSS21pQmM/edit?usp=sharing}} All three spaces are 300-dimensional and their vocabularies are limited to 200K most frequent words.      




\subsection{System Architecture}
\begin{figure*}
    \centering
    \includegraphics[scale=0.4]{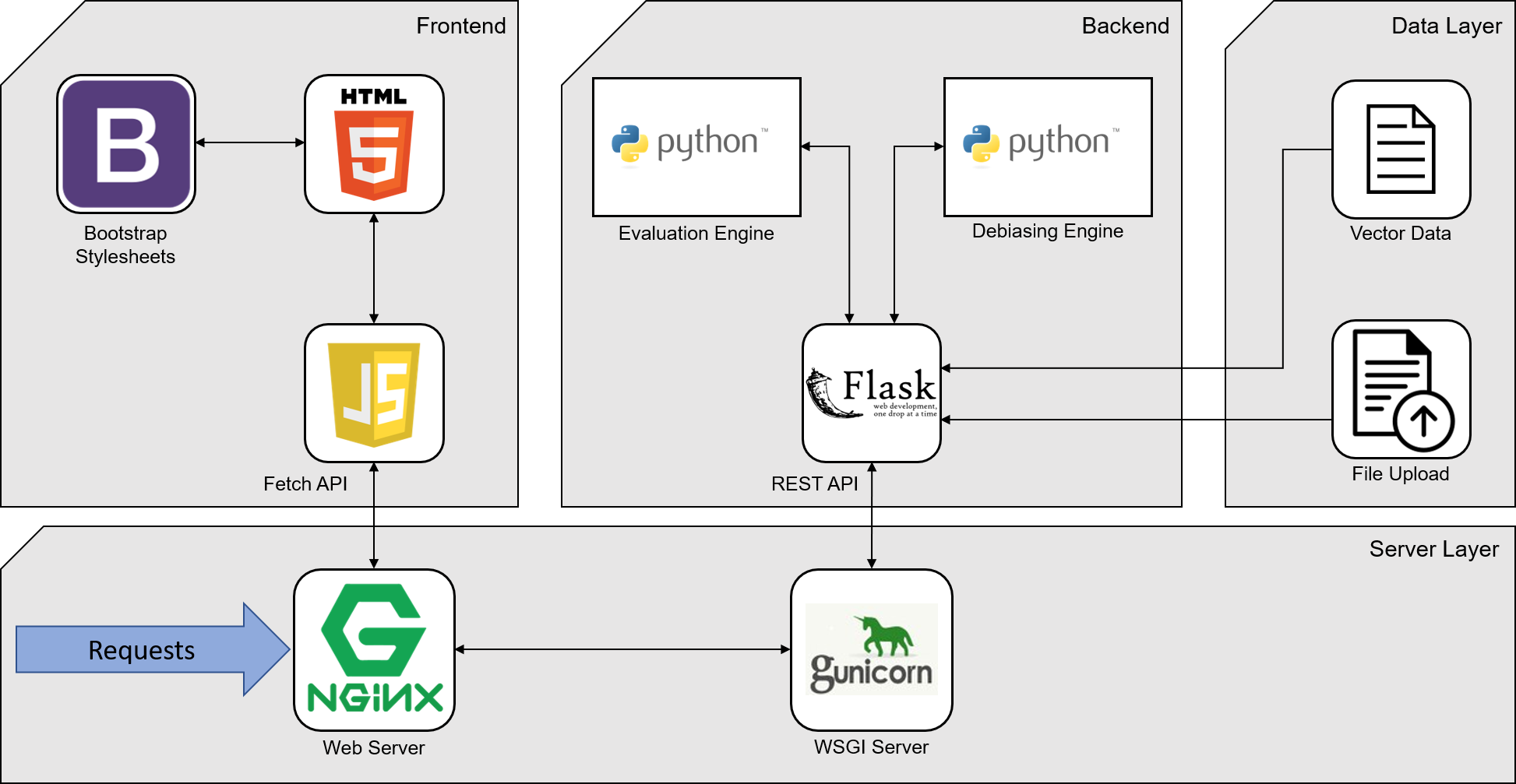}
    \caption{Software Architecture of the \textsc{DebIE} platform.}\label{fig:Architecture_Diagram}
    \vspace{-1em}
\end{figure*}

\textsc{DebIE}'s architecture, illustrated in Figure~\ref{fig:Architecture_Diagram}, adheres to the principles of modern extensible web application design and consists of four components: (1) the backend, (2) the frontend, which together represent the core of the application, (3) the data layer, and (4) the server layer facing the web. 

\paragraph{Backend.} \textsc{DebIE}'s backend consists of two main modules: (1) the bias evaluation engine, which computes the bias test scores (see \S\ref{sec:measures}), and (2) the debiasing engine, which runs the word embedding debiasing models (see \S\ref{sec:debiasing}). The backend interacts with the data layer for retrieving data (bias specifications and vectors from embedding spaces) and its functionality is exposed via a RESTful API, which offers endpoints for programmatically (i) uploading and retrieving data as well as for (ii) running bias evaluation and (iii) debiasing.  
%

There are dedicated controllers and handlers for each of this primary functionalities: vector retrieval, bias evaluation, and debiasing. These are responsible for computing results and delivering content to relevant web pages. The second group of controllers and handlers is responsible for retrieving data out of integrated and external embedding spaces and for parsing and generating JSON data. All bias measures and debiasing methods are implemented as separate modules so that the platform can be extended seamlessly with additional bias measures and debiasing models. A new bias measure or a new debiasing model can be integrated by simply adding the computation scripts (i.e., a function that implements the functionality) and adapting the responsible handler. 
The backend is purely implemented in Python.

\paragraph{Frontend, Data Layer, and Server Layer.} The frontend is written in HTML and plain JavaScript, and relies on the Bootstrap library.\footnote{\url{https://getbootstrap.com/}} The \texttt{fetch} functionality  is used for sending requests to the RESTful API of the backend. For the visualization of embedding spaces (see bottom part of Figure \ref{fig:landing_page}), we rely on the the \texttt{chart.js} library.\footnote{\url{https://www.chartjs.org/}}

Embedding spaces are stored as two files: (1) the \textit{.vocab} file is the serialized dictionary that maps words to indices of the embedding matrix; (2) the \textit{.vectors} file is an embedding matrix (serialized 2D \texttt{numpy} array) rows of which are the actual word vectors. At the start of the web applicatiob, all bias specifications and intergated embedding spaces are fully loaded into the memory completely. 

\textsc{DebIE} is hosted on a Linux server, running Debian 10 as the operating system. The python WSGI-server \texttt{gunicorn} is used to serve the RESTful API. We opt for \texttt{nginx} as the web server for hosting the frontend and redirecting the API-requests to the internal endpoints of the WSGI-server.

\subsection{Accessibility: Interfaces}

Users can interact with \textsc{DebIE} through four different interfaces. The simplest way is by using the provided web interface. For programmatic access, we offer the RESTful-API accessible directly via HTTP requests. 
As a third option, a desktop version of the tool is available for download: this tools runs completely offline and, depending on the hardware, may perform faster. Finally, we offer a command-line interface intended for shell usage.

\paragraph{Web User Interface.} 

\begin{figure}
    \centering
    \includegraphics[scale=0.55]{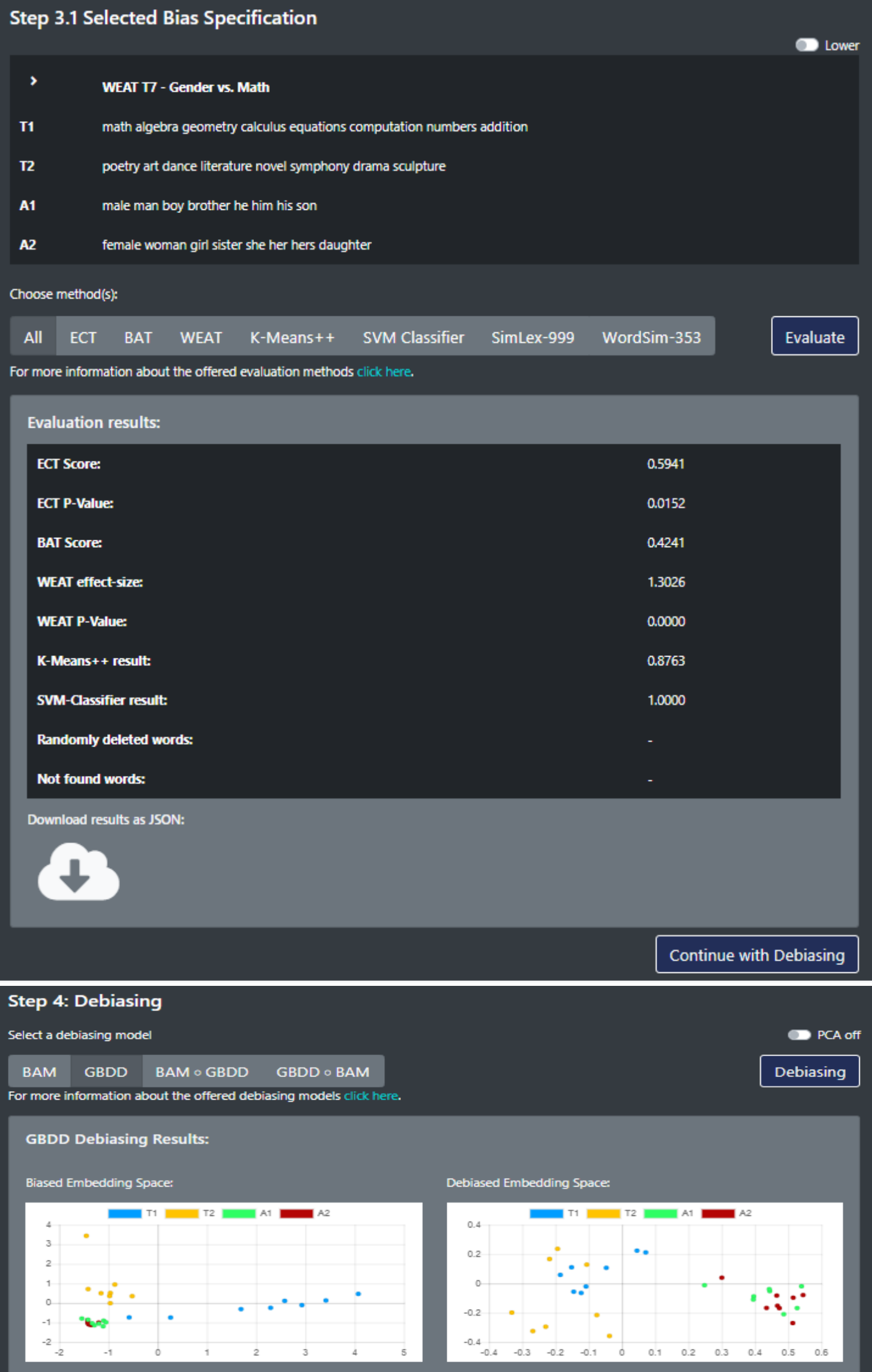}
    \caption{\textsc{DebIE's} web UI.}
    \label{fig:landing_page}
    
    \vspace{-1em}
\end{figure}

\textsc{DebIE} is primarily imagined as a web application with a full extendable web user interface (see Figure~\ref{fig:landing_page}). The web-UI enables users to evaluate and debias with predefined or custom bias specifications. 
Designed as a \textit{one-page application}, the web UI guides the user via five simple steps through the full process:

\noindent\emph{Step 1: Selection of the Embedding Space.} In the first step, the user has to select with which embedding space to work. The users can select one of three integrated embedding spaces (\S\ref{sec:data}) uploaded or their own pretrained vector space. 

\vspace{0.25em}
\noindent\emph{Step 2: Selection of the Bias Specification.}
The user next chooses a bias specification: they can select one of the integrated WEAT bias specifications or define a bias specification of their own. 

\vspace{0.25em}
\noindent\emph{Step 3: Selection and Computation of the Bias Tests.} The user next selects bias measures/scores (see \S\ref{sec:measures}) to be applied on the selected embedding space given the selected bias specification. The bias (and semantic similarity) scores are displayed in a table (see the upper part of Figure \ref{fig:landing_page}) and can also be exported as in the JSON format.

\vspace{0.25em}
\noindent\emph{Step 4: Selection and Execution of Debiasing Algorithms.} 
The user can next choose to debias the selected embedding space (Step 1) based on the selected bias specification (Step 2). To this effect, the user can choose between GBDD, BAM, or one of their compositions (GBDD$\circ$BAM or BAM$\circ$GBDD). The debiased embeddings space can be downloaded. To visualize the differences between the original (biased) and debiased embedding space, we visualize the 2D PCA-compressions of the terms from the bias specification in both spaces (see bottom part of Figure \ref{fig:landing_page}). 

\vspace{0.25em}
\noindent\emph{Step 5: Computation of Bias Tests on the Debiased space.} Finally, the user can evalute the effects of debiasing with the desired set of bias measures. This is like Step 3, only now we subject to testing the debiased instead of the original embedding space.

\paragraph{RESTful API.} 
For programmatic access, we offer a RESTful API. 
The API can deliver vector representations of words, compute and fetch the bias evaluation scores, as well as debiased word embeddings based on a provided bias specification. The API endpoints are accessible online.\footnote{\url{http://debie.informatik.uni-mannheim.de:8000/REST/}} API documentation is available in the \texttt{swagger} format on the \textsc{DebIE} website.\footnote{\url{http://debie.informatik.uni-mannheim.de:8000/swagger/}}

\paragraph{Desktop Application.}
We offer an adapted offline-version of the web application providing the same functionality, runnable on Windows OS. The desktop app has been created with the python module \texttt{flaskwebgui}, using the source files of the web application. The desktop application is available both as a windows executable file (\texttt{.exe}) and as a python script.

\paragraph{Command-line Interface.} 
Finally, we expose \textsc{DebIE}'s functionality through a command-line interface, intended for shell (e.g., \texttt{bash}) usage. We employ the Python framework \texttt{click} to parse the command line arguments.




\section{Ethical Considerations}
Given the high sensitivity of the issue of bias in text representations, we would like the reader to consider the following three aspects. 

(i) Our platform allows for measuring and mitigating biases based on bias specifications, which need to be defined by the user. In actual deployment scenarios, those specifications need to be designed with extreme care and the concrete sociotechnical environment in mind. For instance, it would be wrong to assume that by using one of the predefined gender bias specifications provided with this platform, all stereotypical gender associations will be removed from the representation space. In contrast, for each individual application scenario, the user should make sure that the bias specification matches the bias evaluation and debiasing intent.

(ii) Though the user's main role is to choose appropriate bias specifications, we think it is important that the user has enough technical proficiency to understand potential issues of the provided measures and mitigation methods.

 (iii) The gender bias specifications from previous work provided with this platform only consider bias between \emph{male} and \emph{female} term sets, i.e., they follow a binary notion of gender. However, it is important to keep in mind that gender is a spectrum. We fully acknowledge the importance of the inclusion of \emph{all gender identities}, e.g., nonbinary, gender fluid, polygender, etc., in language technologies.  

\section{Conclusion}
We have presented \textsc{DebIE}, an integrated platform for measuring and attenuating implicit and explicit biases in distributional word vector spaces. Via four different interfaces, we enable fast and easy access to a variety of bias measures and debiasing methods, allowing users to experiment with arbitrary embedding spaces and bias specifications. We hope \textsc{DebIE} facilitates an exploration of a wider set of human biases in language representations. 

\section*{Acknowledgments}
Anne Lauscher and Goran Glavaš are supported by the Eliteprogramm of the Baden-Württemberg Stiftung (AGREE grant). We would like to thank the anonymous reviewers for their helpful comments.

\bibliography{references}
\bibliographystyle{acl_natbib}

\end{document}